\documentclass[]{article}
\usepackage[letterpaper]{geometry}
\usepackage{mtsummit2021}
\usepackage{times}
\usepackage{url}
\usepackage{latexsym}
\usepackage{natbib}
\usepackage{layout}

\usepackage{microtype}
\usepackage{graphicx}
\usepackage{multirow,booktabs}
\usepackage{subcaption}
\usepackage{comment}
\usepackage{xcolor}

\usepackage{microtype}

\begin{document}

\title{ \bf Crosslingual Embeddings are Essential in UNMT for Distant Languages: An English to IndoAryan Case Study}

\author{\name{\bf Tamali Banerjee $\star$} \hfill  \addr{tamali@cse.iitb.ac.in}\\
        \addr{Department of Computer Science and Engineering,
IIT Bombay, India.} \\
\AND
        \name{\bf Rudra Murthy V $\star$} \hfill \addr{rmurthyv@in.ibm.com }\\
        \addr{IBM Research Lab, India.} \\
\AND
       \name{\bf Pushpak Bhattacharyya} \hfill \addr{pb@cse.iitb.ac.in}\\
        \addr{Department of Computer Science and Engineering,
IIT Bombay, India.}
}

\renewcommand{\thefootnote}{\fnsymbol{footnote}}
\footnotetext[1]{The two authors contributed  equally to this paper.}

\maketitle
\pagestyle{empty}

\begin{abstract}

Recent advances in Unsupervised Neural Machine Translation (UNMT) have minimized the gap between supervised and unsupervised machine translation performance for closely related language-pairs. However, the situation is very different for distant language pairs. Lack of lexical overlap and low syntactic similarities such as between English and Indo-Aryan languages leads to poor translation quality in existing UNMT systems.   In this paper, we show that initialising the embedding layer of UNMT models with cross-lingual embeddings shows significant improvements in BLEU score over existing approaches with embeddings randomly initialized.  Further, static embeddings (freezing the embedding layer weights) lead to better gains compared to updating the embedding layer weights during training (non-static). We experimented using Masked Sequence to Sequence (MASS) and Denoising Autoencoder (DAE) UNMT approaches for three distant language pairs. The proposed cross-lingual embedding initialization yields BLEU score improvement of as much as ten times over the baseline for English-Hindi, English-Bengali, and English-Gujarati. Our analysis shows the importance of cross-lingual embedding, comparisons between approaches, and the scope of improvements in these systems.

\end{abstract}

\section{Introduction}

Unsupervised approaches to training a neural machine translation (NMT) system typically involve two stages: (i) Language Model (LM) pre-training and (ii) finetuning of NMT model using Back-Translated (BT) sentences. Training a shared encoder-decoder model on combined monolingual data of multiple languages helps the model learn better cross-lingual representations \citep{conneau-etal-2020-emerging,DBLP:journals/corr/abs-1910-04708}. Fine-tuning the pre-trained model iteratively using Back-translated sentences helps further align the two languages closer in latent space and also trains an NMT system in an unsupervised manner. 


Unsupervised MT has been successful for closely related languages \citep{conneau2019cross,song2019mass}. On the other hand, very poor translation performance has been reported for distant language pairs \citep{kim-etal-2020-unsupervised,marchisio-etal-2020-unsupervised}. Lack of vocabulary overlap and syntactic differences between the source and the target languages make the model fail to align the two language representations together. Recently, few approaches \citep{kulshreshtha2020cross,wu-dredze-2020-explicit} take advantage of resources in the form of bilingual dictionary, parallel corpora, \textit{etc.} to better align the language representations together during LM pre-training.

In this paper, we explore the effect of initialising the embedding layer with cross-lingual embeddings for training UNMT systems for distant languages. We also explore the effect of static cross-lingual embeddings (embedding are not updated during training) \textit{v/s} non-static cross-lingual embeddings (embedding are updated during training). We experiment with two existing UNMT approaches namely, MAsked Sequence-to-Sequence (MASS) \citep{song2019mass} and a variation of Denoising Auto-Encoder (DAE) based UNMT approach \citep{artetxe2018unsupervised,lample2017unsupervised} for English to IndoAryan language pairs  \textit{i.e.} English-Hindi, English-Bengali, English-Gujarati. 

The contribution of the paper is as follows:

\begin{enumerate}
    \item We show that approaches initialized with cross-lingual embeddings significantly outperform approaches with randomly initialized embeddings.
    \item We observe that use of \textit{static cross-lingual embeddings} leads to better gains compared to use of \textit{non-static} cross-lingual embeddings for these language-pairs.
    \item We did a case study of UNMT for English-IndoAryan language pairs. For these language-pairs SOTA UNMT approaches perform very poorly.
    \item We observed that DAE-based UNMT with crosslingual embeddings performs better than MASS-based UNMT  with crosslingual embeddings for these language-pairs.
\end{enumerate}

The rest of the paper is organized as follows. In
Section \ref{sec:related}, we discuss the related work in detail. Then, we present our approach in Section \ref{sec:approach}.In Section \ref{sec:setup}, we outline the experimental setup and present the results of our experiments in Section \ref{sec:results}. Finally, we conclude the paper and discuss future work in Section \ref{sec:conclusion}.

\section{Related Work}
\label{sec:related}

Neural machine translation (NMT)~\citep{cho2014learning,sutskever2014sequence,bahdanau2015neural} typically needs a lot of parallel data to be trained on. However, parallel data is expensive and rare for many language pairs. To solve this problem, unsupervised approaches to train machine translation \citep{artetxe2018iclr, lample2017unsupervised, yang2018unsupervised} was proposed in the literature which uses only monolingual data to train a translation system. 

\cite{artetxe2018unsupervised} and \cite{lample2017unsupervised} introduced Denoising Auto-Encoder-iterative (DAE-iterative) UNMT which utilizes cross-lingual embeddings and trains a RNN-based encoder-decoder model \citep{bahdanau2015neural}. Architecture proposed by \cite{artetxe2018iclr} contains a shared encoder and two language-specific decoders while architecture proposed by \cite{lample2017unsupervised} contains a shared encoder and a shared decoder. In the approach by \cite{lample2017unsupervised}, the training starts with word-by-word translation followed by denoising and backtranslation (BT). Here, noise in the input sentences in the form of shuffling of words and deletion of random words from sentences was performed. 

\cite{conneau2019cross} (XLM) proposed a two-stage approach for training a UNMT system. The pre-training phase involves training of the model on the combined monolingual corpora of the two languages using Masked Language Modelling (MLM) objective \citep{devlin2019bert}. The pre-trained model is later fine-tuned using denoising auto-encoding objective and backtranslated sentences. \cite{song2019mass} proposed a sequence to sequence pre-training strategy. Unlike XLM, the pre-training is performed via MAsked Sequence to Sequence (MASS) objective. Here, random ngrams in the input are masked and the decoder is trained to generate the missing ngrams in the pre-training phase. The pre-trained model is later fine-tuned using backtranslated sentences.

Recently, \cite{kim2020and} demonstrated that the performance of current SOTA UNMT systems is severely affected by language divergence and domain difference. The authors demonstrated that increasing the corpus size does not lead to improved translation performance. The authors hypothesized that existing UNMT approaches fail for distant languages due to lack of mechanism to bootstrap out of a poor initialization.


Recently, \cite{chronopoulou2021improving} trained UNMT systems with 2 language pairs English-Macedonian (En-Mk) and English-Albanian (En-Sq) in low resource settings. These pairs achieved BLEU scores ranging from 23 to 33 using UNMT baseline XLM \citep{conneau2019cross} and RE-LM \citep{chronopoulou2020reusing} systems. They showed further improvement upto 4.5 BLEU score when initialised embedding layer with crosslingual embedding. However, they did not explore the effect of initialising embedding layer on MASS, DAE-pretrained, and DAE-iterative approaches. Moreover, they did not experiment with language-pairs for which UNMT approaches with randomly initialised embedding layer fail completely even after training with a sufficient amount of monolingual data.

Additionally, there is some work on understanding multilingual language models and its effectiveness on zero-shot performance on downstream tasks \citep{pires-etal-2019-multilingual,kulshreshtha2020cross,liu2020multilingual,wang-etal-2020-negative,wu-dredze-2020-explicit}. Here, the pre-trained multilingual language model is fine-tuned for the downstream NLP task in one language and tested on an unseen language (unseen during fine-tuning stage). While multilingual models have shown promising results on zero-shot transfer, the gains are limited for distant languages unless additional resources in the form of dictionary and corpora are used \citep{kulshreshtha2020cross,wu-dredze-2020-explicit}. Also, training a single model on unrelated languages might lead to negative interference \citep{wang-etal-2020-negative}.

\section{Approaches}
\label{sec:approach}
In these section, we explain different approaches used in our experiments. We use MASS \citep{song2019mass} and DAE based iterative approach similar to \cite{lample2017unsupervised} as our baseline models.


\subsection{MASS UNMT} \label{mass}
In MASS \citep{song2019mass}, random n-grams in the input is masked and the model is trained to generate the missing n-grams in the pre-training phase. The pre-trained model is later fine-tuned using back-translated sentences. For every token the input to the model is summation of randomly, initialised word embedding, positional encoding and language code. 

\subsection{DAE UNMT} \label{dae}
DAE UNMT approach is similar the MASS UNMT approach with the difference being the pre-training objective. Here, we add random noise to the input sentence before giving it as input and the model is trained to generate the entire original sentence. Here, noise in the input sentences in the form of shuffling of words and deletion of random words from sentences was performed.

\subsection{Cross-lingual Embedding Initialization}
In both MASS and DAE UNMT approaches, the embedding layer is randomly initialized before the pre-training phase. We use Vecmap \citep{artetxe2018acl} approach as a black-box to obtain cross-lingual embeddings. We then initialize the word-embedding layer with the cross-lingual embeddings obtained. During pre-training and fine-tuning, we have the opportunity to either \textit{freeze} the embedding layer (static embeddings) or update them during training (non-static embeddings). We experiment with both the variations on both MASS and DAE approaches. We refer to MASS UNMT approach using static cross-lingual embeddings as \textit{MASS + Static} and \textit{MASS + Non-Static} for non-static cross-lingual embeddings. Similarly, We refer to DAE UNMT approach using static cross-lingual embeddings as \textit{DAE + Static} and \textit{DAE + Non-Static} for non-static cross-lingual embeddings.

\subsection{DAE-iterative UNMT } \label{undreamtupd}
\cite{artetxe2018unsupervised} and \cite{lample2017unsupervised} proposed an approach based on Denoising Auto-Encoder and Back-Translation. Their approach trained the UNMT in in one stage. During training they alternated between denoising and back translation objectives iteratively. They initialised the embedding layer with cross-lingual embeddings and trained a RNN-based encoder-decoder model \citep{bahdanau2015neural}. Architecture proposed by \cite{artetxe2018iclr} contains a shared encoder and two language-specific decoders while architecture proposed by \cite{lample2017unsupervised} contains a shared encoder and a shared decoder, where all the modules are bi-LSTMs. We use Transformer-based architecture instead of bi-LSTM. In input we do not add language code here. Similar to MASS and DAE, we experiment with using static and non-static cross-lingual embeddings.


\section{Experimental Setup}
\label{sec:setup}
We trained the models using 8 approaches for all language-pair out of which 3 approaches use DAE as LM pretraining, 3 approaches use MASS as LM pretraining, and the other two trains DAE and BT simultaneously.

\subsection{Dataset and Languages used}
We use monolingual data of 4 languages \textit{i.e.} English (en), Hindi (hi),  Bengali (bn), Gujarati (gu). While English is of European language family, the other three languages are of Indo-Aryan language family. These three Indian languages follow Subject-Object-Verb word order. However, for English the word order is Subject-Verb-Object. We organise this experiment for distant language pairs with word-order divergence. Therefore, we pair English language with one of these three Indic languages resulting in three language-pairs, \textit{i.e.} en-hi, en-bn, en-gu.

We use monolingual data provided by AI4Bharat \citep{kunchukuttan2020indicnlpcorpus} dataset as training data. We use Englsh-Indic validation and test data provided in WAT 2020 Shared task \citep{nakazawa2020overview} \footnote{\url{http://www.statmt.org/wmt20/translation-task.html}}. Details of our dataset used in this experiment are in Table \ref{tab:dataset details}.

\begin{table}[!t]
\centering
    \begin{minipage}{0.45\textwidth}
    \centering
    \begin{tabular}{l r }
            \toprule
            \textbf{Language} & \multicolumn{1}{c}{\textbf{\# train}} \\ 
            & \multicolumn{1}{c}{\textbf{ sentences}}\\
            \midrule
            English (en) & 54.3 M\\
            Hindi  (hi) & 63.1 M \\
            Bengali  (bn) & 39.9 M \\
            Gujarati  (gu) & 41.1 M \\
            \bottomrule
    \end{tabular}
    \caption{Monolingual Corpus Statistics in Million}
    \end{minipage}\hfill
    \begin{minipage}{0.45\textwidth}
    \centering
    \begin{tabular}{c | c c}
            \toprule
            \textbf{Language-pair} & \textbf{\# valid} & \textbf{\# test} \\
            & \multicolumn{1}{c}{\textbf{ sentences}} & \multicolumn{1}{c}{\textbf{ sentences}}\\
            \midrule
             En - Hi & 2000 & 3169  \\
             En - Bn & 2000 & 3522 \\
             En - Gu & 2000 & 4463\\
            \bottomrule
    \end{tabular}
    \caption{Validation and Test Data Statistics}
    \label{tab:dataset details}
    \end{minipage}
\end{table}

\subsection{Preprocessing}
We have preprocessed the English corpus for normalization, tokenization, and lowercasing using the scripts available in \textit{Moses} \citep{koehn2007moses} and the Indo-Aryan corpora for tokenization using \textit{Indic NLP Library} \citep{kunchukuttan2020indicnlp}. For BPE segmentation we use \textit{FastBPE}\footnote{\url{https://github.com/glample/fastBPE}} jointly on the source and target data with number of merge operations set to 100k.

\subsection{Word Embeddings}
We use the BPE-segmented monolingual corpora to independently train the embeddings for each language using skip-gram model of \textit{Fasttext}\footnote{\url{https://github.com/facebookresearch/fastText}}  \citep{bojanowski2017enriching}. To map embeddings of the two languages to a shared space, we use \textit{Vecmap}\footnote{\url{https://github.com/artetxem/vecmap}} to obtain cross-lingual embedding proposed by \citet{artetxe2018robust}. We report the quality of the cross-lingual embeddings in Table \ref{tab:emb_accuracy} w.r.t. word-translation quality on MUSE data \citep{conneau2017word} by nearest-neighbour and Cross-Domain Similarity Local Scaling (CSLS) approaches. 

\begin{table}[!t]
\centering
    \begin{tabular}{c | c c c c}
            \toprule
            \multirow{2}{*}{\textbf{Language-pair}} &  \multicolumn{2}{c}{\textbf{en $\rightarrow$ x}} &  \multicolumn{2}{c}{\textbf{x $\rightarrow$ en}} \\
            & NN & CSLS & NN & CSLS \\
            \midrule
            
            En - Hi & 52.16 \% & 55.46 \% & 43.51 \% & 46.82 \% \\
            En - Bn & 36.76 \% & 41.39 \% & 33.77 \% & 39.17 \% \\
            En - Gu & 43.35	\% & 46.47	\% & 46.07 \%	 & 50.38 \% \\
            \bottomrule
            \end{tabular}
        \caption{Word-to-word translation accuracy using our crosslingual embeddings}

    \label{tab:emb_accuracy}
\end{table}

\subsection{Network Parameters}
We use MASS code-base \footnote{\url{https://github.com/microsoft/MASS}} and to tun our experiments. We train all the models with a $6$ layer $8$-headed transformer encoder-decoder architecture of dimension 1024. The model is trained using an epoch size of $0.2M$ steps and a batch-size of 64 sentences (token per batch $3K$)). We use Adam optimizer with $beta_1$ set to 0.9, and $beta_2$ to 0.98, with learning rate to 0.0001. We pre-training for a total of 100 epochs and fine-tune for a maximum of 50. However, we stop the training if the model converges before the max-epoch is reached. The input to the model is a summation of word embedding and positional encoding of dimension 1024. In all our models, we drop the language code at the encoder side. For MASS pre-training we use word-mass of $0.5$. Other parameters are default parameters given in the code-base.

\subsection{Evaluation and Analysis}
We report both BLEU scores as translation accuracy metric for these approaches. We additionally plot perplexity, accuracy, and BLEU scores for intermediate results of each model.

\section{Result and Analysis}
\label{sec:results}
In this section, we present the results from our experiments and present a detailed analysis of the same.

\subsection{Results}
The translation performance from our experiments is as shown in Table \ref{result}. We compared BLEU scores between models where embedding layers were initialised with cross-lingual embeddings and models where embedding layers were randomly initialised.

Initialising embedding layer with static cross-lingual embedding helps both MASS-based and DAE-based UNMT systems to learn better translations as seen from the table. Our results suggest that freezing cross-lingual embeddings (static) during UNMT training results in better translation quality compared to the approach where cross-lingual embeddings are updated (non-static).

\begin{figure*}
    \centering
    \includegraphics[width=\textwidth]{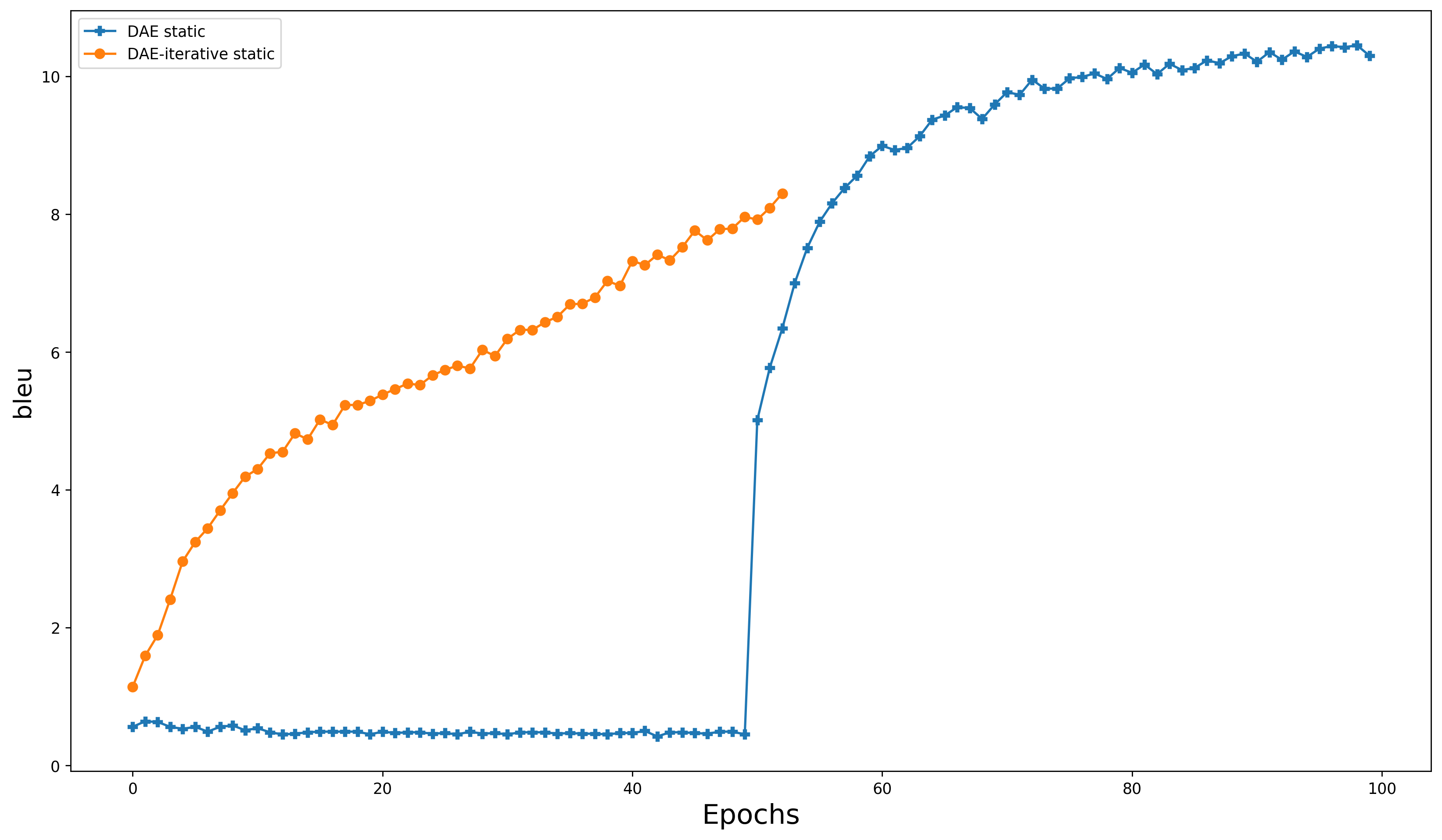}
    \caption{Comparison of Test Set BLEU Score for every epoch between DAE Static (DAE-pretrained UNMT) (both pre-training and fine-tuning) and DAE-iterative Static approach. Embedding layers of both the approaches are initialised with cross-lingual embedding and frozen during training. Language-pair: English-Hindi.}
    \label{enhiundbleu}
\end{figure*}

BLEU scores suggest that DAE objective based models surpass MASS objective based models for these language pairs. Though DAE-iterative models produce lower BLEU scores than \textit{DAE Static} or \textit{DAE Non-Static} models, the former approach gives better BLEU scores in less number of iterations as shown in Fig. \ref{enhiundbleu}. 

For completeness, we compare the BLEU scores of the best UNMT model, \textit{i.e.} \textit{DAE Static}, with the best reported BLEU scores in WAT 2020 Shared Task \citep{nakazawa2020overview} reported by \cite{yu2020hw} on the same test data in the supervised setting. The supervised approach uses parallel data in a multilingual setting. Their models reached high accuracy by improving baseline multilingual NMT models with Fast-align, Domain transfer, ensemble, and Adapter fine-tuning methods. 

While our en-hi and en-gu models produce decent values of BLEU score, en-bn models produce low BLEU score. Intuitively, we assume language characteristics to be the reason behind it.



    
    

\begin{table*}[ht!]
    \centering
     \resizebox{\textwidth}{!}{ 
    \begin{tabular}{l rr rr rr}
    \toprule
    \multirow{1}{*}{\textbf{UNMT approaches}} & \multirow{1}{*}{en $\rightarrow$ hi} & \multirow{1}{*}{hi $\rightarrow$ en} & \multirow{1}{*}{en $\rightarrow$ bn} & \multirow{1}{*}{bn $\rightarrow$ en} & \multirow{1}{*}{en $\rightarrow$ gu} & \multirow{1}{*}{gu $\rightarrow$ en} \\
    \midrule
     MASS & 1.15 & 1.61 & 0.11 & 0.27 & 0.62 & 0.79 \\
    DAE & 0.63 &	0.95 &	0.06 &	0.31 &	0.39 &	0.61	\\
    \midrule
    DAE-iterative Non-Static & 5.37 &	6.63 &	1.66 &	4.19 &	3.12 &	5.98 \\
    MASS Non-Static & 5.49 &	6.06 &	1.86 &	3.5	& 3.47 &	4.82	\\
    DAE Non-Static & 7.65&	8.85&	2.35&	4.67&	4.55&	6.84 \\
    \midrule
    DAE-iterative Static & 7.96 &	9.09 &	2.88 &	5.54 &	5.63 &	8.64 \\
    MASS Static & 5.5&	6.49&	2.09&	4.7&	4.13&	6.09 \\
    DAE Static & \textbf{10.3} &	\textbf{11.57} & \textbf{3.3} &	\textbf{6.91} &	\textbf{7.39} &	\textbf{10.88} 	\\\bottomrule
    \end{tabular}
    }
    \caption{UNMT translation performance on distant languages, i.e. en-hi, en-bn, en-gu test sets (BLEU scores reported). The values marked in bold indicate the best score for a language pair.}
    \label{result}
    
\end{table*}

\begin{table}[ht!]
    \centering
     \resizebox{\textwidth}{!}{ 
    \begin{tabular}{l rr rr rr}
    \toprule
     \textbf{System} & \multirow{2}{*}{en $\rightarrow$ hi} & \multirow{2}{*}{hi $\rightarrow$ en} & \multirow{2}{*}{en $\rightarrow$ bn} & \multirow{2}{*}{bn $\rightarrow$ en} & \multirow{2}{*}{en $\rightarrow$ gu} & \multirow{2}{*}{gu $\rightarrow$ en} \\
    &  & &  &  &  \\
    \midrule
     
    Our best UNMT & 10.3 &	11.57 & 3.3 &	6.91 &	7.39 &	10.88 	\\
    
    \midrule
    
    SOTA Supervised NMT & 24.48 &	28.51 & 19.24 &	23.38 &	14.16 &	30.26 	\\
    
    \bottomrule
    
    \end{tabular}
    }
    \caption{Comparison of results between our best unsupervised NMT models and SOTA supervised NMT models on WAT20 test data. Supervised NMT results are reported from \cite{yu2020hw}}.
    \label{supunsupresults}
    
\end{table}

\begin{figure}[!htb]
    \centering
    \includegraphics[width=0.8\textwidth]{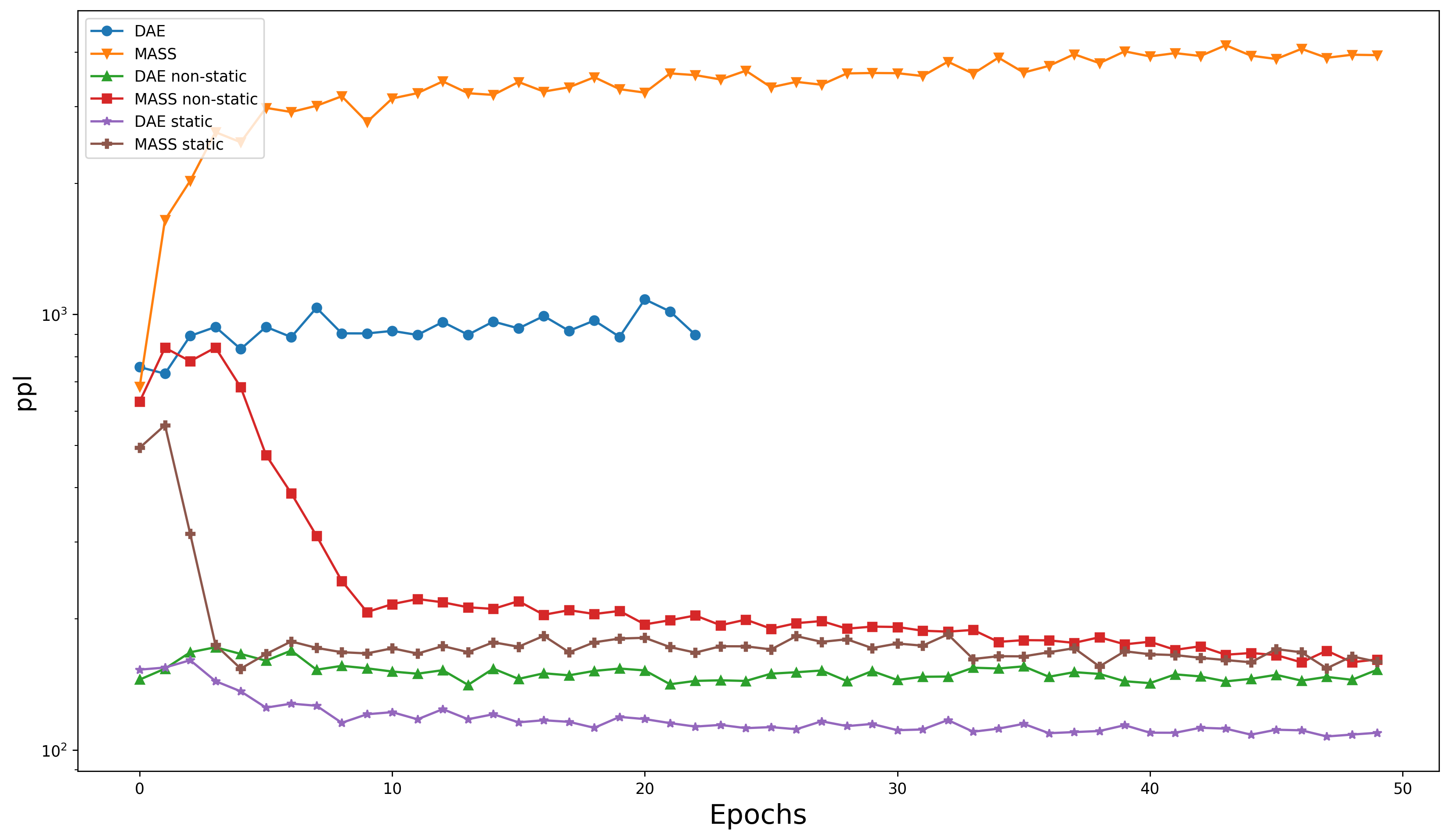} 
    \caption{Change in Validation Set Translation Perplexity during fine-tuning for English to Hindi Language pair}
    \label{enhiplotft}
\end{figure}

\begin{figure}

    \centering
    \includegraphics[width=0.7\textwidth]{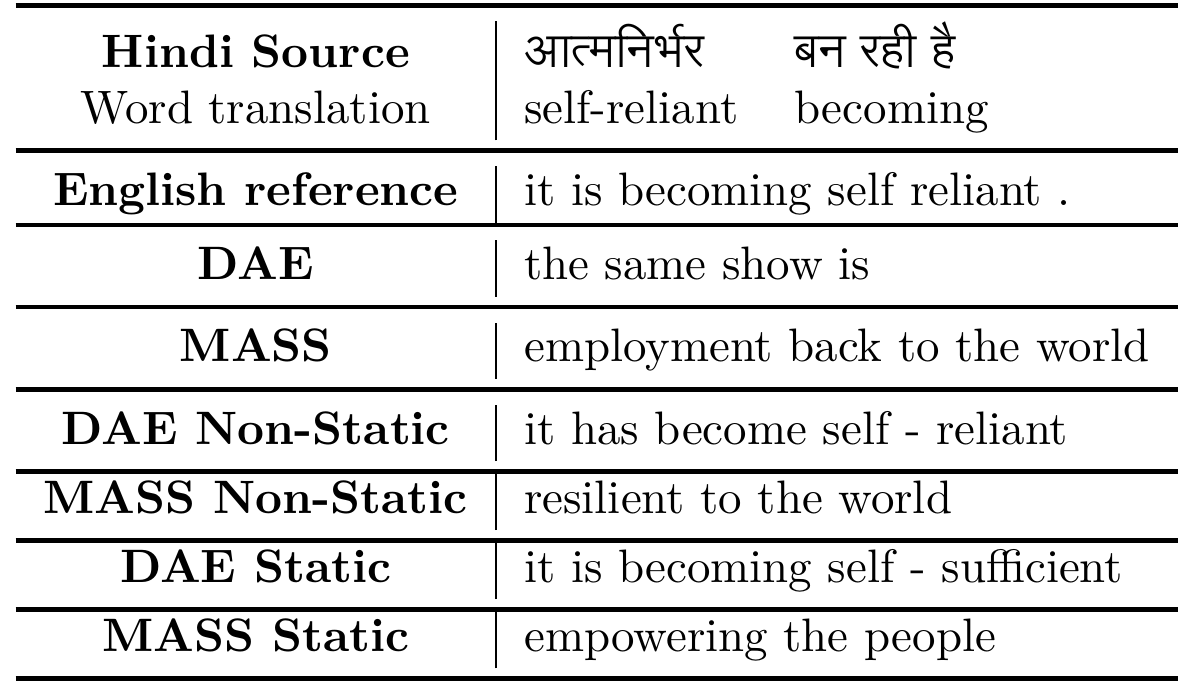}
    \caption{Example of a Hindi to English translation using various approaches}
    \label{examplehien}
\end{figure}

\subsection{Analysis}
We analyse the performance of our models by plotting translation perplexities on the validation set. Moreover, we manually analyse translation outputs and discuss them in this section.

\subsubsection{Quantitative Analysis}

In Fig. \ref{enhiplotft}, we observe that for both MASS (baseline MASS) and DAE (baseline DAE) the plot of translation perplexity over epoch increases rather than decreasing. On the other hand, when cross-lingual word embeddings are used the validation set translation perplexity decreases.

Among these embedding initialised models, we observe better convergence for models where embedding layers are frozen (static) than the models where embedding layers are updated (non-static). We also observe that the DAE-UNMT models converge better than MASS-UNMT models when initialized with cross-lingual embeddings.



\subsubsection{Qualitative Analysis}
An example of a Hindi $\rightarrow$ English translation produced by various approaches is presented in Fig. \ref{examplehien}. We observe the translation to be capturing the meaning of the source sentence when cross-lingual embeddings are used. However, we report some observations we found while analysing the translation outputs.

\begin{figure*}
    \centering
    \includegraphics[width=\textwidth]{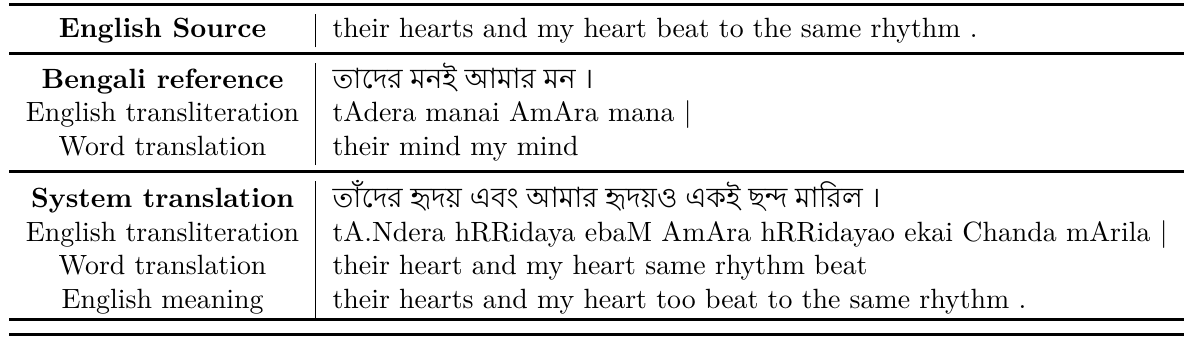}
    \caption{Example of a English to Bengali translation using DAE Static model}
    \label{enbnliteral}
\end{figure*}

\paragraph{Lose of Phrasal Meaning} We observe some translations where word meanings are prioritised over phrasal meaning. Fig. \ref{enbnliteral} shows such an example where dis-fluent translation is generated because of ignoring the phrasal meaning. Here, the model is unable to get the conceptual meaning of the sentence, instead translates words of the sentence literally. 

\begin{figure*}
    \centering
    \includegraphics[width=\textwidth]{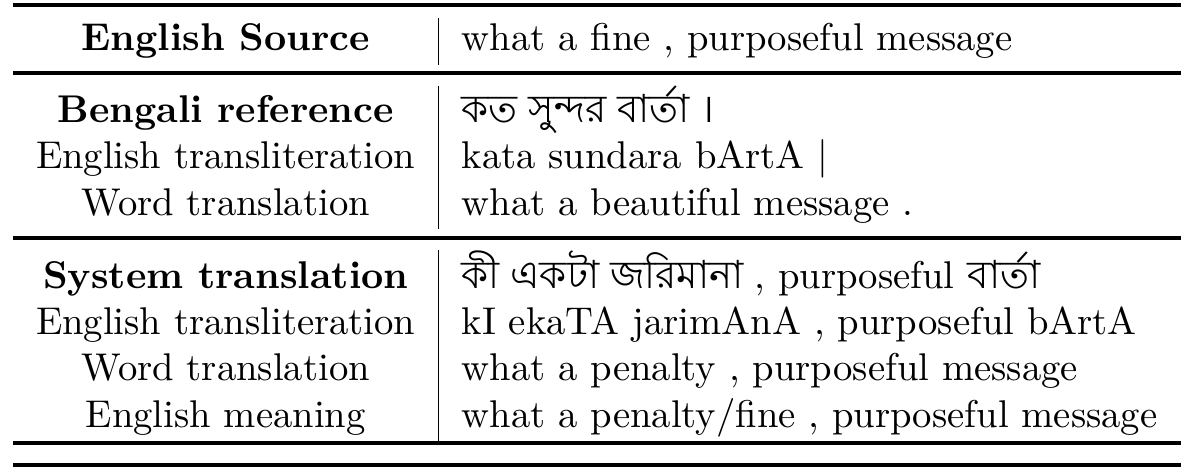}
    \caption{Example of a English to Bengali translation using DAE Static model}
    \label{enbnsense}
\end{figure*}

\paragraph{Word Sense Ambiguity} In Fig. \ref{enbnsense} model fails to disambiguate word sense resulting in wrong translation. English word \textit{`fine'} have different sense, \textit{i.e.} beautiful and penalty. In this example, the model selects wrong sense of the word.

\begin{figure*}
    \centering
    \includegraphics[width=\textwidth]{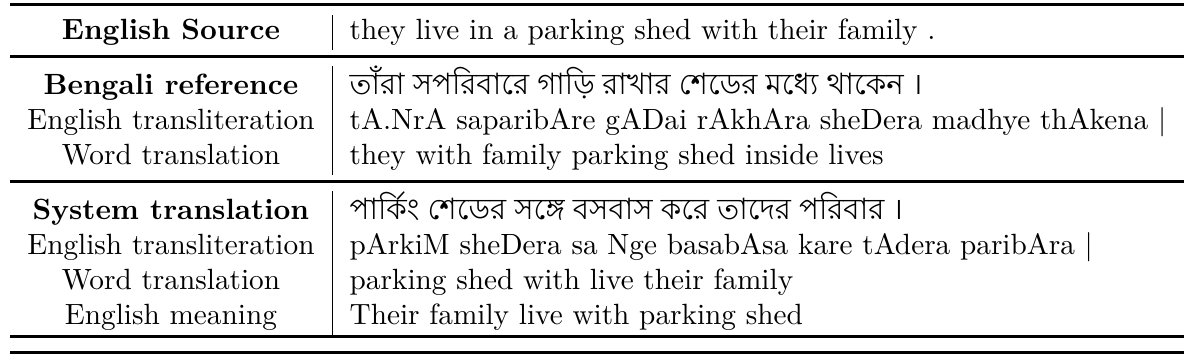}
    \caption{Example of a English to Bengali translation using DAE Static model}
    \label{enbnscramble}
\end{figure*}

\paragraph{Scrambled Translation} For many instances like Fig. \ref{enbnscramble}, though the reference sentence and its corresponding generated sentences are formed with almost the same set of words, the sequence of words is different making the sentence lose its meaning.




\section{Conclusion}
\label{sec:conclusion}
We show that existing UNMT methods such as DAE-based and MASS-based UNMT models fail for distant languages such as English to IndoAryan language pairs (\textit{i.e.} en-hi, en-bn, en-gu). However, initialising the embedding layer with cross-lingual embeddings before Language Model (LM) pre-training helps the model train better UNMT systems for distant language pairs. We also observe that static cross-lingual embedding gives better translation quality compared to non-static cross-lingual embeddings. For these distant language pairs, DAE objective based UNMT approaches produce better translation quality and converges better than MASS-based UNMT. 

\bibliographystyle{apalike}
\bibliography{mtsummit2021}


\end{document}